\documentclass{article} % For LaTeX2e
\usepackage{iclr2027_conference,times}

\usepackage{amsmath,amsfonts,bm}

\newcommand{\E}[1]{\mathbb{E}\left[#1\right]}
\newcommand{\defined}{\stackrel{def}{=}}
\newcommand{\range}[1]{r\!\left(#1\right)}

\newcommand{\quant}[1]{\widetilde{#1}}
\newcommand{\quantf}[1]{Q\left({#1}\right)}
\newcommand{\diag}[1]{\text{Diag}\left(#1\right)}
\newcommand{\loss}[1]{{\mathcal{L}\left(#1\right)}}
\newcommand{\vect}[1]{{\text{vec}\left(#1\right)}}

\newcommand{\trace}[1]{\text{Tr}\left(#1\right)}
\newcommand{\eye}{\mI}

\def\eqref#1{equation~\ref{#1}}
\def\1{\bm{1}}

\def\vg{{\bm{g}}}

\def\vs{{\bm{s}}}

\def\vv{{\bm{v}}}
\def\vw{{\bm{w}}}
\def\vx{{\bm{x}}}

\def\mA{{\bm{A}}}
\def\mB{{\bm{B}}}

\def\mD{{\bm{D}}}

\def\mF{{\bm{F}}}
\def\mG{{\bm{G}}}
\def\mH{{\bm{H}}}
\def\mI{{\bm{I}}}

\def\mL{{\bm{L}}}
\def\mM{{\bm{M}}}

\def\mP{{\bm{P}}}
\def\mQ{{\bm{Q}}}
\def\mR{{\bm{R}}}

\def\mT{{\bm{T}}}
\def\mU{{\bm{U}}}

\def\mW{{\bm{W}}}
\def\mX{{\bm{X}}}

\def\mZ{{\bm{Z}}}

\DeclareMathAlphabet{\mathsfit}{\encodingdefault}{\sfdefault}{m}{sl}
\SetMathAlphabet{\mathsfit}{bold}{\encodingdefault}{\sfdefault}{bx}{n}

\newcommand{\KL}[2]{{D_{\mathrm{KL}}\left(#1 || #2\right)}}
\newcommand{\D}[2]{{D\left(#1 || #2\right)}}

\newcommand{\Cov}{\mathrm{Cov}}
\DeclareMathOperator*{\argmin}{arg\,min}

\usepackage{hyperref}
\usepackage{url}
\usepackage{graphicx} 
\usepackage{amsmath}
\usepackage{amssymb}
\usepackage[table]{xcolor}
\usepackage{booktabs}
\usepackage{algorithm} \usepackage{algpseudocode}
\usepackage{multirow}
\iclrfinalcopy

\colorlet{BlockColor}{yellow!20}

\title{A Unified Hessian-Guided Framework for Post-Training Quantization}

\author{ 
    \vspace{-0.3cm}~\\ 
    \bf{Mads Toftrup\thanks{Equal contribution (shared first authorship).}~~\!\thanks{Work done while doing an internship at Qualcomm AI Research.}}\\ 
    Aarhus University\\ 
    \texttt{toftrup@cs.au.dk} \And 
    \vspace{-0.3cm}~\\ 
    \bf{Marco Federici$^{*}$}\\ 
    Qualcomm AI Research\\ 
    \texttt{mfederic@qualcomm.com} \And 
    \vspace{-0.3cm}~\\ 
    \bf{Boris van Breugel}\\ 
    Qualcomm AI Research\\ 
    \texttt{bvanbreu@qualcomm.com} 
    \AND 
    \vspace{-0.7cm}~\\ 
    \bf{Paul Whatmough}\\ 
    Qualcomm AI Research\\ 
    \texttt{pwhatmou@qualcomm.com} \And 
    \vspace{-0.7cm}~\\ 
    \bf{Markus Nagel}\\ 
    Qualcomm AI Research\\ 
    \texttt{markusn@qualcomm.com} 
}

\begin{document}

\maketitle

\begin{abstract}
Low-bit quantization enables efficient deployment of large language models by reducing memory footprint and computational cost. Achieving high accuracy at low precision requires selecting quantization ranges, linear transformations, and weight rounding decisions. Existing approaches typically optimize these components independently using local reconstruction objectives, which fail to account for the varying impact of quantization errors on downstream predictions. Direct optimization of task-level metrics would provide a more faithful objective, but is challenging due to the discontinuous nature of quantization operations and the high computational cost of evaluating their downstream effects.

In this work, we derive a Hessian-guided approximation of the output change induced by joint weight and activation quantization. Based on a tractable Fisher approximation, we formulate a unified post-training quantization framework that enables the joint optimization of quantization ranges, linear transformations, and weight rounding under a common output-sensitivity objective. The resulting method remains practical for modern large language models and consistently outperforms state-of-the-art transformation and rounding methods across both channel-wise and block-wise quantization settings.
\end{abstract}

\begingroup
\renewcommand\thefootnote{}% Remove the marker
\footnotetext{Qualcomm AI Research is an initiative of Qualcomm Technologies, Inc.}
\endgroup

\section{Introduction}

Large language models achieve strong performance across a wide range of language tasks, but their memory and computational requirements make deployment expensive. Quantization reduces these costs by representing weights and activations at lower precision, improving memory efficiency and enabling faster integer computation. The central challenge is to obtain these benefits while preserving the predictions of the full-precision model.

We focus on integer post-training quantization (PTQ) of both weights and activations. A small calibration set informs the quantization parameters. At a fixed bit width, accuracy depends on the quantization ranges, rounding choices, and representation basis. Quantization ranges set the scales: narrower ranges give finer grids but can increase clipping error. Rounding then selects weights on those grids. Both choices operate on a given representation of the layer. An invertible, function-preserving transform~\citep{FPTQuant}, such as a rotation or Hadamard transform \citep{ashkboos_quarot_2024}, changes that representation by redistributing weight and activation magnitudes while preserving the full-precision layer output. Range selection and transforms can address both weights and activations, whereas adaptive rounding primarily addresses weights.

\begin{figure}[!tbp]
    \centering
    \includegraphics[width=\linewidth]{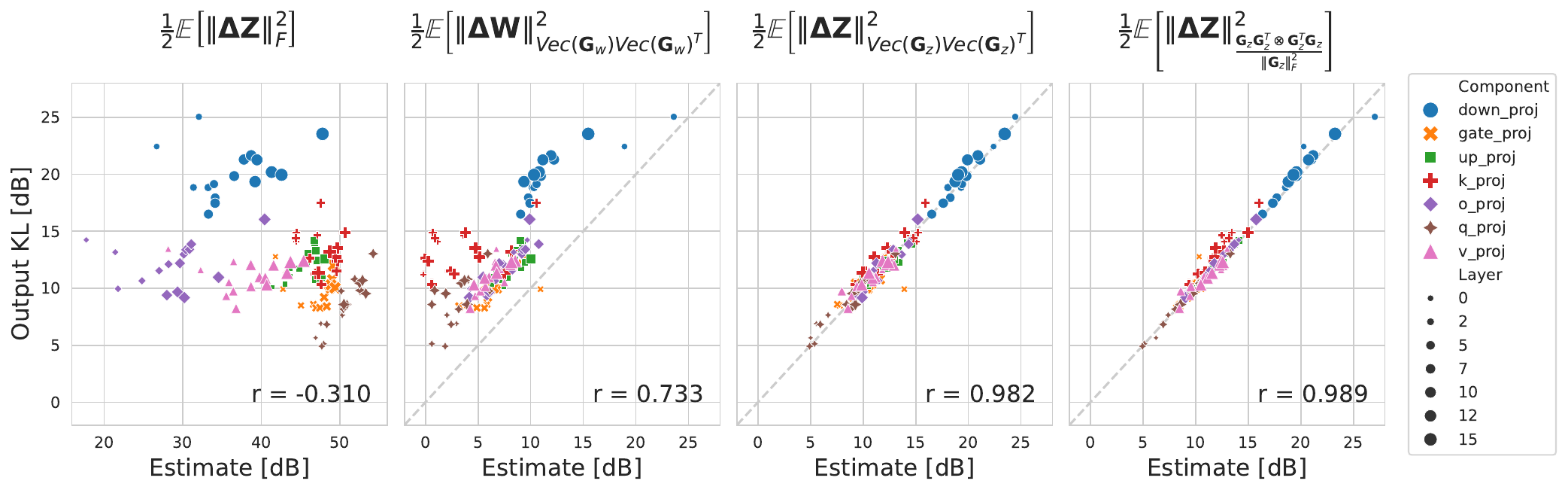}
    \vspace{-0.7cm}
    \caption{Relationship between the KL divergence between the original and quantized model outputs and several layer-local error proxies across linear layers of Llama 3.2 1B Instruct under W4A4 quantization. Conventional reconstruction metrics, such as the output Frobenius norm, exhibit weak correlation with downstream output degradation. Fisher-weighted weight error fails to capture the effect of activation quantization. Output Fisher-weighted error correlates more strongly with the KL divergence, and our proposed Kronecker factorization of the Output Fisher  retains this correlation.}
    \label{fig:figone}
\end{figure}

Local reconstruction evaluates these choices by the error they introduce at a layer's output. But, as shown in Figure~\ref{fig:figone}, the magnitude of local perturbations is not necessarily a good proxy for the changes in the model output. Existing methods~\citep{Tseng25} capture these downstream effects through an approximation to the Hessian of the full-model output KL divergence with respect to a layer's weights. However, methods based on the Weight Hessian fail to capture the error introduced by quantizing the activations (Figure~\ref{fig:figone}, second column).

In this work, we use a quadratic expansion of the KL divergence between the original and quantized model predictions around the layer output with a tractable approximation (Figure~\ref{fig:figone}, third and last column). This lets us derive a unified algorithm that optimizes transformations, range setting, and weight rounding under the same curvature metric, accounting for the downstream effects of both weight \emph{and} activation quantization.

\begin{samepage}
Our contributions are:
\begin{itemize}
    \item We introduce a layer Output Kronecker-factorized  approximation to the layer-output Fisher matrix that defines a quadratic proxy for the KL divergence between original and quantized outputs, accounting for both weight \textit{and} activation quantization.
    \item Using the output Kronecker factorization, we derive $\mH_\mZ$-PTQ, a unified procedure for transformation design, weight range selection, adaptive rounding, and activation clipping under a common downstream-sensitivity objective.
    \item We demonstrate improvements over existing PTQ baselines across multiple model families and quantization granularities.
\end{itemize}

\end{samepage}

\section{Background}
\label{sec:background}
We first introduce the quantization choices considered in this work. 
Let $\mX\in\mathbb{R}^{s\times i}$ represent the activation matrix for $s$ tokens and $i$ input features, and let $\mW\in\mathbb{R}^{i\times o}$ be the weights of a linear layer with $o$ output features. Its output is $\mZ=\mX\mW$.

\paragraph{Quantization scales and clipping.}
For signed integer quantization with $b\geq 2$ bits, $k_{\min}=-2^{b-1}$ and $k_{\max}=2^{b-1}-1$. 
Given a grid step $s>0$ and shift $\mu$, round-to-nearest quantization is
\begin{equation}
    Q(a;s,\mu)
    =\mu+s\operatorname{clip}\!\left(
    \operatorname{round}\!\left(\frac{a-\mu}{s}\right),
    k_{\min},k_{\max}\right).
    \label{eq:background_uniform_quantizer}
\end{equation}
The grid step $s$ determines the tradeoff between clipping and rounding resolution. Increasing $s$ expands the representable range and reduces clipping, but produces a coarser quantization grid. Decreasing $s$ improves resolution within the representable range, but may clip large-magnitude values.

Unlike weights, whose grid steps $\vs_w$ are fixed after calibration, activation scales depend on the current input. We therefore optimize a shared clipping factor $\gamma_x \in (0,1]$~\citep{sun_flatquant_2024} and compute the corresponding input-dependent activation steps $\vs_x(\mX,\gamma_x)$ at runtime. The activations $\vx_i$ are clipped to $[-\gamma_x \max_i |\vx_i|,\gamma_x \max_i |\vx_i|]$ under symmetric quantization and to $[\gamma_x (\min_i \vx_i-\mu),\gamma_x (\max \vx_i-\mu)]$ under asymmetric quantization. We then apply Eq.~\ref{eq:background_uniform_quantizer} using a grid fitted to these bounds, with a zero-centered grid in the symmetric case and a shifted grid in the asymmetric case.

\paragraph{Weight rounding.}

Once the representation and quantization grid are fixed, rounding assigns each weight to a representable value. Round-to-nearest treats weights independently, whereas adaptive rounding methods coordinate these decisions so that later choices can compensate for errors introduced earlier~\citep{AdaRound,GPTQ}. Existing methods generally optimize either a layer-reconstruction objective or a weight-space approximation to the downstream model loss.

\paragraph{Transformations.}
For any invertible transformation $\mT\in\mathbb{R}^{i\times i}$, transforming the activations and weights preserves the full-precision output, $(\mX\mT)(\mT^{-1}\mW)=\mZ$, but generally changes the quantized output, $\quant{\mZ}=Q(\mX\mT)Q(\mT^{-1}\mW)$. We write $\Delta\mZ=\mZ-\quant{\mZ}$ for the resulting layer-output perturbation. If $\mT$ reduces a weight group's range under the same quantizer, $Q(\mT^{-1}\mW)$ can use a finer grid and have lower weight error than $Q(\mW)$ in their respective representations~\citep{sun_flatquant_2024}.
Diagonal transforms scale channels~\citep{Nagel_2019_ICCV,SmoothQuant,AWQ}. Orthogonal transforms mix channels through randomized Hadamard transforms~\citep{QuIPSharp,QuaRot}, learned rotations~\citep{SpinQuant}, or blockwise rotations with permutations~\citep{DuQuant}. More general transforms combine rotations and scaling~\citep{OSTQuant}. FlatQuant jointly learns invertible Kronecker factors, channel scaling, and clipping through block reconstruction~\citep{sun_flatquant_2024}. CAT~\citep{Federici2026} and WUSH~\citep{Chen2026} design transforms to minimize a local error proxy.

\section{Method}

Our derivation proceeds in four steps. We first express PTQ as the minimization of the prediction divergence induced by replacing one full-precision layer with its quantized counterpart. We then approximate this divergence by a quadratic form in the layer-output perturbation. Because the resulting output Hessian is intractably large, we introduce a tractable approximation. Finally, we use this common quadratic proxy to derive practical optimization procedures for transformations, weight scales and rounding, and activation clipping.

\subsection{Joint objective}

Ideally, post-training quantization would jointly select the transforms, weight and activation grids, and  weight rounding decisions to minimize the downstream discrepancy between the full-precision and quantized models. For a linear layer, this idealized objective can be written as:
\begin{align}
    \min_{\gamma_x,\vs_w,\mT,\mW^*}
    \E{\loss{
        \underbrace{
            \overbrace{Q(\mX\mT;\vs_x(\mX\mT,\gamma_x))}^{\text{quantized activations } \quant{\mX}}
            \overbrace{Q(\mT^{-1}\mW^*;\vs_w)}^{\text{quantized weights } \quant{\mW}}
        }_{\text{quantized output } \quant{\mZ}}
    }},
    \label{eq:goal_abstract}
\end{align}
in which $\loss{}$ refers to a measure of change of the model output.

To model the downstream effect of weight and activation quantization, we use a second-order Taylor expansion around the full-precision layer output $\mZ$.
\begin{align}
    \loss{\quant{\mZ}} \approx \loss{\mZ} + \vect{\Delta\mZ}^T\vect{\mG_{\mZ}} + \frac{1}{2}\vect{\Delta\mZ}^T\mH_{\mZ}\vect{\Delta\mZ},
    \label{loss:step_0}
\end{align}
in which $\vect{}$ refers to a vectorization operation that stacks the rows of a matrix into a single column vector, $\mG_{\mZ}$ is the loss gradient with respect to $\mZ$, and $\mH_{\mZ}$ is the (flattened) Hessian matrix with shape $[s\times$o, s$\times$o].

Common choices of loss $\loss{}$ include measures of divergence of the model output distribution with respect to the original (unperturbed model). For these choices of loss, the zero and first order terms vanish since $\mZ=\quant\mZ$ is a global minimum for which the divergence is zero: 
\begin{align}
    \D{p_\mZ}{p_{\quant{\mZ}}} \approx \frac{1}{2}\vect{\Delta\mZ}^T\mH_{\mZ}\vect{\Delta\mZ}\defined \frac{1}{2}\|\Delta\mZ\|_{\mH_{\mZ}}^2 ,
    \label{loss:step_1}
\end{align}

where $p_\mZ$ and $p_{\quant{\mZ}}$ refer to the model output distribution induced by the original and quantized layer, respectively.
 We use the shorter expression $\|\Delta\mZ\|_{\mH_{\mZ}}^2$ to indicate that the norm of the layer output perturbation is weighted by a quadratic proxy of the output loss sensitivity $\mH_{\mZ}$.

\subsection{Modeling the Hessian matrices}

Empirically estimating a full rank Hessian matrix is intractable for modern-sized deep learning architectures due to the computation and memory requirements.
However, when selecting the Kullback-Leibler divergence for Equation~\ref{loss:step_1}, the Hessian becomes equivalent to the Fisher Information Matrix (FIM).
The (empirical) Fisher matrix can be estimated by considering the outer product of the flattened Output gradients with respect to sampled labels from the output distribution: 
\begin{align}
    \mH_{\mZ} = \mF_{\mZ}\defined\E{\vect{\mG_{\mZ}}\vect{\mG_{\mZ}}^T}.
\end{align}

Estimating a full Fisher matrix, which is quadratic in the product of sequence length and output feature dimension, is clearly still not practical for full-scale deep neural networks. For this reason, following a similar procedure as \citep{Tseng25}, we simplify the computation by factorizing the effect of the FIM on the sequence dimension $ \mF_{s(\mZ)}$ and feature output dimension $\mF_{f(\mZ)}$:
\begin{align}
    \mF_{\mZ} \approx \mF_{\otimes\mZ}\defined 
    \frac{\overbrace{\mG_{\mZ}\mG_{\mZ}^T}^{\mF_{s(\mZ)}}\otimes\overbrace{\mG_{\mZ}^T\mG_{\mZ}}^{\mF_{f(\mZ)}}}{\|\mG_{\mZ}\|_F^2}
    ,
    \label{eq:output_kronecker}
\end{align}
in which $\mF_{s(\mZ)}$ and $\mF_{f(\mZ)}$ represent proxies for the sequence and feature curvatures, respectively. 
% \mn{do we have an appendix with the derivation or assumptions here?}

We refer to this approximation as the \textbf{Output Kronecker} Fisher factorization, which exactly reproduces each sampled gradient outer product when $\mG_{\mZ}$ is rank 1.
A summary and comparison with the literature is included in Table~\ref{tab:hessian_approx}, while the correlation between the Output Kronecker Fisher weighted error and the KL divergence is visualized in Figure~\ref{fig:figone}.

In order to perform the optimization, we estimate and store the following quantities on a small calibration set:  the expected output feature curvature  $\E{\mG_\mZ^T\mG_\mZ}$, the weight input curvature $\E{\mG_\mW\mG_\mW^T}$ and the token gradient magnitude vector $\E{\diag{\mG_\mZ\mG_\mZ^T}}$. The input feature curvature can be obtained at runtime by left and right multiplying the output feature curvature by the weight matrices $\E{\mG_\mX^T\mG_\mX}=\mW\E{\mG_\mZ^T\mG_\mZ}\mW^T$.

\begin{table}[!th]
    \centering
    \caption{Comparison of Fisher Information Metric estimates for linear layer output, weights and input implied by different approaches in literature.  Normalization factors are omitted for brevity.}
    \footnotesize
    \setlength{\tabcolsep}{3pt}
    \begin{tabular}{l|r@{$\;\otimes\;$}l|r@{$\;\otimes\;$}l|r@{$\;\otimes\;$}l|l}
    \toprule
          Estimate &  \multicolumn{2}{c|}{$\overbrace{\mF_{s(\mZ)}\otimes \mF_{f(\mZ)}}^{\text{Output Fisher: }\mF_{\otimes\mZ}}$~~~~} & 
          \multicolumn{2}{c|}{$\overbrace{\mF_{i(\mW)}\otimes \mF_{o(\mW)}}^{\text{Weight Fisher: }\mF_{\otimes\mW}}$} & 
          \multicolumn{2}{c|}{$\overbrace{\mF_{s(\mX)}\otimes \mF_{f(\mX)}}^{\text{Input Fisher: }\mF_{\otimes\mX}}$~~~~~~~~~~~~~~~~}
          &  Assumptions \\\midrule
         Isotropic & $\eye_s$ & $ \eye_o$ & $\mX^T\mX $ & $ \eye_o$ & $  \eye_s $ & $ \mW\mW^T $
         & Isotropic  $\mF_\mZ$ \\
         K-FAC & $\eye_s $ & $ \mG_{\mZ}^T\mG_{\mZ}$ & $\mX^T\mX $ & $ \mG_{\mZ}^T\mG_{\mZ}$ & $\eye_s $ & $ \mG_{\mX}^T\mG_{\mX}$
         & Isotropic  $\mF_{s(\mZ)}$ \\
         YAQA & $\mG_\mZ\mG_\mZ^T $ & $ \mG_{\mW}^T\mG_{\mW}$ & $\mG_{\mW}\mG_{\mW}^T $ & $ \mG_{\mW}^T\mG_{\mW}$ & $\mG_\mZ\mG_\mZ^T $ & $ \mW\mG_{\mW}^T\mG_{\mW}\mW^T$
         & Rank-1 $\mG_\mW$ \\
         \textbf{Ours} & $\mG_\mZ\mG_\mZ^T $ & $ \mG_{\mZ}^T\mG_{\mZ}$ & $\mG_{\mW}\mG_{\mW}^T $ & $ \mG_{\mZ}^T\mG_{\mZ}$ & $\mG_\mZ\mG_\mZ^T $ & $ \mG_{\mX}^T\mG_{\mX}$
         & Rank-1 $\mG_\mZ$ \\
         \cline{1-8}
         \bottomrule
    \end{tabular}
    \label{tab:hessian_approx}
\end{table}

\subsection{\texorpdfstring{$\mH_\mZ$-PTQ}{HZ-PTQ}: Optimization Procedure}

We start optimizing the output-Fisher-weighted proxy by considering the linear transformation $\mT$. This is because the transformation $\mT$ affects model weights, activations, and intermediate gradients, making a joint optimization very challenging:
\begin{align}
    \min_{\gamma_x,\vs_w,\mT,\mW^*_\mT} \KL{p_\mZ}{p_{\quant{\mZ}}}\approx \frac{1}{2}\min_{\gamma_x,\vs_w,\mT,\mW^*_\mT}\E{\|\Delta\mZ\|_{\mF_{\otimes\mZ}}^2}\approx \frac{1}{2}\min_{\gamma_x,\vs_w,\mW^*_\mT}\E{\|\Delta\mZ_{\hat{\mT}}\|_{\mF_{\otimes\mZ}}^2},
\end{align}
in which $\hat{\mT}$ represents the optimal transformation $\arg\min_\mT\E{\|\Delta\mZ\|_{\mF_{\otimes\mZ}}^2}$, and $\Delta\mZ_{\hat{\mT}}$ is the output discrepancy with the transformation $\hat{\mT}$ applied.
Once $\hat{\mT}$ is fixed, we denote the transformed activations
and weights by $\mX_{\hat{\mT}}=\mX\hat{\mT}$ and
$\mW_{\hat{\mT}}=\hat{\mT}^{-1}\mW$, respectively.
We express the corresponding activation and weight Fisher
estimates in these coordinates and use them to determine
the remaining quantization parameters.

Once the transformations are fixed, using mild independence assumptions, we can factorize the effect of activation and weight quantization: 
\begin{align}
    \E{\|\Delta\mZ_{\hat{\mT}}\|_{\mF_{\otimes\mZ}}^2} 
    \approx \underbrace{\E{\|\mX_{\hat{\mT}}\overbrace{(\mW_{\hat{\mT}}-\quant{\mW}_{\hat{\mT}})}^{\Delta\mW_{\hat{\mT}}}\|_{\mF_{\otimes\mZ}}^2}}_{\text{Weight quantization error}}+\underbrace{\E{ \|\overbrace{(\mX_{\hat{\mT}}-\quant{\mX_{\hat{\mT}}})}^{\Delta\mX_{\hat{\mT}}}\mW_{\hat{\mT}}\|_{\mF_{\otimes\mZ}}^2}}_{\text{Activation quantization error}}.
    \label{loss:step_2}
\end{align}
The approximate split in Eq.~\ref{loss:step_2} permits disentangling weight- and activation-specific parameters.
First, we find the optimal weight scale and rounding:
\begin{align}
    \hat{\vs_\mW}, \hat{\mW}_{\hat{\mT}}^* = \arg\min_{\vs_w,\mW^*}\E{\|\mX_{\hat{\mT}}\Delta\mW_{\hat{\mT}}\|_{\mF_{\otimes\mZ}}^2}.
\end{align}
Lastly, we can determine the optimal clipping based on the updated weights $\hat{\mW}_{\hat{\mT}}^*$ by determining:
\begin{align}
    \hat{\gamma_x} = \arg\min_{\gamma_x}\E{\|\Delta\mX_{\hat{\mT}}\hat{\mW}_{\hat{\mT}}^*\|_{\mF_{\otimes\mZ}}^2}.
\end{align}
We refer to the resulting framework as $\mH_\mZ$-PTQ because all optimization stages are derived from the same layer-output curvature proxy $\mF_{\otimes\mZ}$. Sections~\ref{sec:transforms}--\ref{sec:clip} describe each stage.

\subsection{Optimal Transformations}
\label{sec:transforms}
In order to find the optimal transformation $\hat{\mT}$, we first use
 standard rounding-noise assumptions to explicitly express our quadratic loss proxy as a function of the transformed quantization ranges for activation and weights, which determine the stretching of the quantization grid, and the norm of the transformed gradients:
\begingroup
\small
\begin{equation}
    \E{\|\Delta\mZ_\mT\|_{\mF_{\otimes\mZ}}^2}\approx\frac{\E{\sum_t\alpha_t\range{\vx_t\mT}^2}}{12(2^{b_x}-1)^2}\E{\|\mG_{\mX}\mT^{-T}\|_F^2}+\frac{\E{\sum_j\beta_j\range{\mT^{-1}\vw_j}^2}}{12(2^{b_w}-1)^2}\E{\|\mT^T\mG_{\mW}\|_F^2}
    \label{loss:approx_full}
\end{equation}
\endgroup
The activation and weight bit widths are $b_x,b_w$, and $\range{\vv}$ denotes the range of the quantization grid.
The weights $\alpha_t=\|(\mG_{\mZ})_{t,:}\|_2^2/\|\mG_{\mZ}\|_F^2$ and $\beta_j=\|(\mG_{\mZ})_{:,j}\|_2^2/\|\mG_{\mZ}\|_F^2$ measure the relative gradient energy of token $t$ and output feature $j$.
They remain unchanged by the transforms because $\mX_{\mT}\mW_{\mT}=\mZ$.
A full derivation is reported in Appendix~\ref{loss_decomposition_derivations}.
Overall, as expected, the expression in Eq.~\ref{loss:approx_full} is invariant to scaling of $\mT$. Furthermore, the norm of the transformed gradients $\|\mG_{\mX}\mT^{-T}\|_F^2$ and $\|\mT^{T}\mG_{\mW}\|_F^2$ are invariant to orthogonal transformations. 

Following \citet{Federici2026}, we approximately minimize Eq. \ref{loss:approx_full} with a two-step procedure.
First, we choose the symmetric positive-definite component $\mM$ to minimize the sum of squared transformed gradient norms:
\begin{equation}
     \hat{\mM} = \argmin_{\mM\succ0}
     \E{\|\mG_{\mX}\mM^{-T}\|_F^2}
     + \E{\|\mM^T\mG_{\mW}\|_F^2}
     = \left(
         \E{\mG_{\mW}\mG_{\mW}^T}^{-1}
         \mathbin{\#}
         \E{\mG_{\mX}^T\mG_{\mX}}
     \right)^{1/2}.
     \label{eq:gradient_balance}
\end{equation}
Here, $\#$ denotes the matrix geometric mean, see Appendix~\ref{app:gradient_balance} for the derivation.
In practice, we use a block-wise approximation $\hat{\mM}_{\text{block}}=\text{BlockDiag}(\hat{\mM},\text{block})$ to reduce the memory and computation cost \citep{Federici2026,Chen2026}.

Secondly, we apply a Hadamard rotation to reduce the quantization ranges \citep{Federici2026}.
The final transformation is obtained by combining the two components $\hat{\mT}=\hat{\mM}_{\text{block}}\mH$.

Note that by making the additional assumption that the output gradients are isotropic ($\mG_\mZ\propto\eye$), the expression in Eq.~\ref{loss:approx_full} simplifies to the one reported in \citet{Federici2026}, and the solution in Eq.~\ref{eq:gradient_balance} is equivalent to the transforms proposed in \citet{Federici2026, chen2026wush}, for which $\hat{\mM}$ aligns the directions of variations of activations and weights instead of their gradients.

\subsection{Weight scales and rounding}
\label{sec:range_and_round}

The optimal quantization scales and rounding can be determined by considering:
\begin{align}
\label{weight_range_eq}
    \min_{\vs_w,\mW^*_{\hat{\mT}}}\E{\|\mX_{\hat{\mT}}\Delta\mW_{\hat{\mT}}\|_{\mF_{\otimes\mZ}}^2}
    &\approx
    \min_{\vs_w,\mW^*_{\hat{\mT}}}
    \left\|
        \mW_{\hat{\mT}}-\quantf{\mW^*_{\hat{\mT}};\vs_w}
    \right\|_{
        \frac{\hat{\mT}^T\E{\mG_{\mW}\mG_{\mW}^T}\hat{\mT}
        \otimes
        \E{\mG_{\mZ}^T\mG_{\mZ}}}{\E{\|\mG_\mZ\|^2_2}}
    }^2.
\end{align}

In which  $\hat{\mT}^T\E{\mG_{\mW}\mG_{\mW}^T}\hat{\mT}$ determines the (transformed) weight input-side curvature while $\E{\mG_{\mZ}^T\mG_{\mZ}}$ the output-side curvature. In order to disentangle the optimization of scales and rounding, we follow a procedure similar to the one introduced in \citet{Tseng25}:
first we perform a block-LDL decomposition of the input and output curvature based on the size of the quantization tile $[k_i,k_o]$. Here $k_i$ represents the number of input channels that share the same scale, while $k_o$ represents the number of output channels that share the same scale (usually 1):
\begin{align}
    \mL_{i}, \mD_{i}=\text{BlockLDL}\left(\hat{\mT}^T\E{\mG_{\mW}\mG_{\mW}^T}\hat{\mT}; k_i\right), \mL_{o}, \mD_{o}=\text{BlockLDL}\left(\E{\mG_{\mZ}^T\mG_{\mZ}}; k_o\right).
\end{align}
Then, we use an iterative procedure in which we alternately set a quantization scale and adjust the weight rounding, starting from the original (transformed) weights $\mW^{(0)}=\mW_{\hat{\mT}}$.

The curvature for the quantization tile in coordinate $(j,k)$ is  fully determined by the block diagonal components of $\mD_i[j]$ and $\mD_o[k]$. Since $\mD_o[k]$ is a scalar, the optimal scale $s_{jk}$ can be determined by selecting:
\begin{align}
    \hat{s}_{jk}=\arg\min_{s_{jk}}\left\|\mW_{\hat{\mT},B(j,k)}-\quantf{\mW^{(0)}_{B(j,k)};s_{jk}}\right\|_{\mD_i[j]}^2,
    \label{eq:min_d}
\end{align}
in which $B(j,k)$ is used to indicate the entries of the block $(j,k)$.
The expression in Eq.~\ref{eq:min_d} can be easily optimized using a grid search over a few candidates to pick the best scale.
Fixing the scale $\hat{s}_{jk}$, we can determine the quantization error given by the quantization function in the block $(j,k)$: $\Delta\mW^{(0)}_{B(j,k)}=\mW_{\hat{\mT},B(j,k)}-\quantf{\mW^{(0)}_{B(j,k)};\hat{s}_{jk}}$ and zero everywhere else.
Lastly, we can use the $\mL_i$ and $\mL_o$ matrices to determine the correction and update the rounded weight for the remaining blocks 
\begin{align}
    \mW^{(1)} =  \mW^{(0)} + \mL_i^T\Delta\mW^{(0)}\mL_o
\end{align}
Proceeding in anti-diagonal order starting from the bottom-right, the quantization scales can be set one at a time, and the final $\mW^{(i/k_i\times o/k_o)}$ represents the adjusted weights $\hat{\mW}^*_{\hat{\mT}}$. The key difference from \citet{Tseng25} lies in using $\mD_i$ to set the scale of each block and in the different Fisher proxy. Similar to \citet{frantar_gptq_2022}, before running the algorithm, we sort the block rows and columns so that blocks corresponding to larger Fisher traces are quantized first.

\subsection{Activation clipping.}
\label{sec:clip}
For the activation clipping it is non-trivial to determine a good
$\gamma_x \in (0,1]$: a heuristic constant value may not be optimal, but training it end-to-end is expensive and may yield sub-optimal solutions. We
propose to use the Hessian proxy as a good middle ground that takes into account the downstream sensitivity.
\begin{align}
    \E{\|\Delta\mX_{\hat{\mT}}\mW_{\hat{\mT}}\|_{\mF_{\otimes\mZ}}^2}
    &\approx\E{\|\mX_{\hat{\mT}}-Q(\mX_{\hat{\mT}};\vs_x(\mX_{\hat{\mT}},\gamma_x))\|_{\E{\diag{\mG_\mZ\mG_\mZ^T}}\otimes\hat{\mT}^{-1}\E{\mG_{\mX}^T\mG_{\mX}}\hat{\mT}^{-T}}^2},
\end{align}
The optimal $\gamma_x$ can then be determined on a small calibration set by picking the minimum error over a small set of candidates. Note that instead of using the dense sequence curvature $\E{\mG_\mZ\mG_\mZ^T}$, we consider only a diagonal component $\E{\diag{\mG_\mZ\mG_\mZ^T}}$. This is to drastically reduce the memory requirements since only one vector needs to be stored. This approximation results in little error since we empirically validated that the sequence curvature component is mostly diagonal. 

\section{Experimental Results}

We evaluate our downstream-output proxy, compare quantization performance across models and quantization granularities, and ablate the contributions of individual components.
Appendix~\ref{app:experimental_details} experimental details and Appendix~\ref{app:additional_results} includes additional results.

\paragraph{Quantization and Calibration} We consider two quantization granularities and evaluate three configurations: (i) \textbf{W4A4KV4}, a channel-wise setting in which weights, activations, and the KV cache are quantized to 4-bit integers~\citep{sun_flatquant_2024,ashkboos_quarot_2024,Federici2026}; weights use one scale per output channel, while activations use one scale per token; (ii) \textbf{W4-g32}, a block-wise weight-only setting in which groups of 32 weight values share a floating-point scale, similar to the one used in \citet{Tseng25}; and (iii) \textbf{W4A4-g32}, a block-wise weight-and-activation setting in which groups of 32 values share a floating-point scale~\citep{chen2026wush}. For all methods, quantization parameters and Hessian estimates are computed using 2048 randomly sampled sequences of 2048 tokens from the DCLM-Edu dataset~\citep{dclmedu}.

\paragraph{Evaluation Protocol} Following standard evaluation practice for LLM quantization \citep{sun_flatquant_2024, Federici2026}, we report perplexity on WikiText \citep{merity_pointer_2017} using a context length of 2048 tokens and the mean normalized accuracy across six common-sense reasoning tasks (csr):  PIQA~\citep{bisk_piqa_2020}, WinoGrande~\citep{sakaguchi_winogrande_2021}, HellaSwag~\citep{zellers_hellaswag_2019}, ARC-e and ARC-c~\citep{clark_think_2018}, and LAMBADA~\citep{paperno_lambada_2016}.
For methods involving calibration-dependent optimization, we report the mean and standard deviation over three random calibration seeds.

\begin{table}[!t]
    \centering
    \caption{W4A4KV4 channel-wise quantization results. The effectiveness of $\mH_\mZ$-PTQ  demonstrates the advantage of the joint optimization under a unified Hessian-guided objective.}
    \setlength{\tabcolsep}{1.5pt}
    \scriptsize
    \begin{tabular}{l|ll|ll|ll|ll} \toprule & \multicolumn{2}{c|}{\textit{Llama v3.2 1B inst.}} & \multicolumn{2}{c|}{\textit{Llama v3 8B}} & \multicolumn{2}{c|}{\textit{Ministral 8B inst.}} & \multicolumn{2}{c}{\textit{Qwen v3 8B}} \\ Method & wiki ($\downarrow$) & csr ($\uparrow$) & wiki ($\downarrow$) & csr ($\uparrow$)& wiki ($\downarrow$) & csr ($\uparrow$)& wiki ($\downarrow$) & csr ($\uparrow$)\\ \midrule 
FP & 13.16 & 57.61 & 6.14 & 70.12 & 6.97 & 71.29 & 9.73 & 67.46 \\\midrule
RTN & 878.89 & 30.02 & 317.60 & 32.40 & 39.73 & 40.62 & 10441.69 & 29.23 \\
QuaRot$^{+\text{GPTQ}}$ & ${20.03}^{\pm 0.33}$ & ${48.93}^{\pm 0.56}$ & ${8.34}^{\pm 0.04}$ & ${63.39}^{\pm 0.15}$ & ${8.18}^{\pm 0.00}$ & ${67.77}^{\pm 0.09}$ & ${34.43}^{\pm 9.39}$ & ${51.64}^{\pm 3.05}$ \\
CAT$^{+\text{GPTQ}}$ & ${17.68}^{\pm 0.09}$ & ${52.19}^{\pm 0.46}$ & ${7.53}^{\pm 0.02}$ & ${65.38}^{\pm 0.13}$ & ${7.90}^{\pm 0.02}$ & ${68.06}^{\pm 0.21}$ & ${10.67}^{\pm 0.02}$ & ${64.60}^{\pm 0.34}$ \\
FlatQuant$^{+\text{GPTQ}}$ & ${17.23}^{\pm 0.11}$ & ${52.95}^{\pm 0.40}$ & ${7.19}^{\pm 0.01}$ & $\bf{67.32}^{\pm 0.55}$ & ${7.82}^{\pm 0.01}$ & ${68.88}^{\pm 0.33}$ & ${10.60}^{\pm 0.03}$ & ${65.26}^{\pm 0.11}$ \\
$\mH_\mZ$-PTQ & $\bf{16.05}^{\pm 0.02}$ & $\bf{54.21}^{\pm 0.11}$ & $\bf{7.06}^{\pm 0.01}$ & $\bf{67.25}^{\pm 0.23}$ & $\bf{7.69}^{\pm 0.01}$ & $\bf{69.32}^{\pm 0.37}$ & $\bf{10.43}^{\pm 0.03}$ & $\bf{65.56}^{\pm 0.46}$ \\

\cline{1-9}\bottomrule \end{tabular}
    \label{tab:table_1}
\end{table}

\subsection{Channel-wise W4A4KV4 Quantization}
We first evaluate the W4A4KV4  setting, where only a single scale is available for each weight output channel and activation token. Results are reported in Table~\ref{tab:table_1}.
$\mH_\mZ$-PTQ consistently achieves the lowest perplexity across all evaluated architectures while matching or improving reasoning accuracy. Importantly, the gains are observed for every model family considered, ranging from the 1B-parameter Llama model to larger 8B-parameter architectures. This consistency indicates that the benefits of the proposed objective are not tied to a specific model architecture or size.
Compared to FlatQuant, which learns quantization-aware transformations through layer-wise optimization, $\mH_\mZ$-PTQ achieves lower perplexity on all four models. Likewise, $\mH_\mZ$-PTQ consistently outperforms CAT, despite both methods relying on the same transformation with different parameters. 

\begin{table}[!t]
    \centering
    \scriptsize
    \caption{For block-wise quantization, $\mH_\mZ$-PTQ consistently achieves the lowest perplexity across both W4-g32 and W4A4-g32 settings, demonstrating that the proposed Hessian-guided objective remains effective even at finer quantization granularities.}
    \setlength{\tabcolsep}{2pt}\begin{tabular}{ll|ll|ll|ll|ll}
\toprule
 &  & \multicolumn{2}{c|}{\textit{Llama v3.2 1B inst.}} & \multicolumn{2}{c|}{\textit{Llama v3 8B}} & \multicolumn{2}{c|}{\textit{Ministral 8B inst.}} & \multicolumn{2}{c}{\textit{Qwen v3 8B}} \\
  & Method & wiki ($\downarrow$) & csr ($\uparrow$) & wiki ($\downarrow$) & csr ($\uparrow$) & wiki ($\downarrow$) & csr ($\uparrow$) & wiki ($\downarrow$) & csr ($\uparrow$) \\
\midrule
&FP & 13.16 & 57.61 & 6.14 & 70.12 & 6.97 & 71.29 & 9.73 & 67.46 \\
\midrule
\multirow[m]{6}{*}{\rotatebox[origin=c]{90}{W4-g32}} & RTN & $15.47$ & $55.27$ & $6.65$ & $68.67$ & $7.28$ & $70.88$ & $9.99$ & $65.95$ \\
 & Had.$^{+\text{GPTQ}}$ & ${13.73}^{\pm 0.02}$ & ${56.83}^{\pm 0.18}$ & ${6.43}^{\pm 0.01}$ & ${68.97}^{\pm 0.19}$ & ${7.13}^{\pm 0.01}$ & ${71.01}^{\pm 0.12}$ & ${9.99}^{\pm 0.05}$ & ${66.80}^{\pm 0.15}$ \\
 & CAT$^{+\text{GPTQ}}$ & ${13.77}^{\pm 0.08}$ & ${56.77}^{\pm 0.17}$ & ${6.44}^{\pm 0.01}$ & ${69.64}^{\pm 0.18}$ & ${7.14}^{\pm 0.00}$ & $\bf{70.99}^{\pm 0.03}$ & ${9.92}^{\pm 0.04}$ & ${67.14}^{\pm 0.28}$ \\
 & WUSH$^{+\text{GPTQ}}$ & ${13.85}^{\pm 0.08}$ & ${56.81}^{\pm 0.14}$ & ${6.44}^{\pm 0.01}$ & ${69.43}^{\pm 0.33}$ & ${7.13}^{\pm 0.01}$ & ${70.83}^{\pm 0.05}$ & ${9.88}^{\pm 0.08}$ & ${67.06}^{\pm 0.32}$ \\
 & Had.$^{+\text{YAQA}}$ & ${13.74}^{\pm 0.02}$ & $\bf{57.02}^{\pm 0.18}$ & ${6.39}^{\pm 0.00}$ & ${69.61}^{\pm 0.03}$ & ${7.22}^{\pm 0.00}$ & $\bf{70.96}^{\pm 0.12}$ & ${9.93}^{\pm 0.06}$ & ${67.29}^{\pm 0.22}$ \\
 & $\mH_\mZ$-PTQ & $\bf{13.53}^{\pm 0.04}$ & $\bf{57.04}^{\pm 0.19}$ & $\bf{6.32}^{\pm 0.00}$ & $\bf{69.93}^{\pm 0.17}$ & $\bf{7.09}^{\pm 0.01}$ & $\bf{70.97}^{\pm 0.04}$ & $\bf{9.84}^{\pm 0.01}$ & $\bf{67.42}^{\pm 0.25}$ \\
\midrule
\multirow[m]{6}{*}{\rotatebox[origin=c]{90}{W4A4-g32}} & RTN & $17.99$ & $51.75$ & $7.25$ & $66.68$ & $7.72$ & $69.21$ & $10.39$ & $63.89$ \\
 & Had.$^{+\text{GPTQ}}$ & ${14.57}^{\pm 0.02}$ & ${55.59}^{\pm 0.29}$ & ${6.72}^{\pm 0.01}$ & ${67.66}^{\pm 0.37}$ & ${7.35}^{\pm 0.02}$ & ${70.19}^{\pm 0.23}$ & ${10.19}^{\pm 0.06}$ & ${66.20}^{\pm 0.27}$ \\
 & CAT$^{+\text{GPTQ}}$ & ${14.54}^{\pm 0.09}$ & ${55.64}^{\pm 0.45}$ & ${6.71}^{\pm 0.01}$ & ${68.66}^{\pm 0.20}$ & ${7.40}^{\pm 0.01}$ & ${70.02}^{\pm 0.14}$ & ${10.11}^{\pm 0.04}$ & ${66.38}^{\pm 0.11}$ \\
 & WUSH$^{+\text{GPTQ}}$ & ${14.65}^{\pm 0.04}$ & ${55.28}^{\pm 0.45}$ & ${6.74}^{\pm 0.02}$ & ${67.98}^{\pm 0.54}$ & ${7.42}^{\pm 0.01}$ & ${69.83}^{\pm 0.21}$ & ${10.11}^{\pm 0.09}$ & ${66.30}^{\pm 0.31}$ \\
 & Had.$^{+\text{YAQA}}$ & ${14.56}^{\pm 0.02}$ & ${55.53}^{\pm 0.17}$ & ${6.68}^{\pm 0.01}$ & ${68.51}^{\pm 0.15}$ & ${7.46}^{\pm 0.01}$ & ${70.10}^{\pm 0.11}$ & ${10.13}^{\pm 0.07}$ & ${66.22}^{\pm 0.33}$ \\
 & $\mH_\mZ$-PTQ & $\bf{14.22}^{\pm 0.05}$ & $\bf{55.75}^{\pm 0.16}$ & $\bf{6.56}^{\pm 0.00}$ & $\bf{68.82}^{\pm 0.15}$ & $\bf{7.29}^{\pm 0.00}$ & $\bf{70.38}^{\pm 0.13}$ & $\bf{9.99}^{\pm 0.01}$ & $\bf{66.77}^{\pm 0.38}$ \\
\cline{1-10}
\bottomrule
\end{tabular}
    \label{tab:table_2}
\end{table}

\subsection{Block-wise Quantization}
We next consider block-wise quantization using groups of size 32. This setting is particularly challenging for transformation-based methods since smaller quantization groups already reduce range variability, leaving less room for improvement through range equalization alone.
Following \citet{chen2026wush}, we apply independent block-wise transforms before quantization and fuse the corresponding inverses into the weights. We evaluate both weight-only (W4-g32) and weight-and-activation (W4A4-g32) quantization scenarios. Results are reported in Table~\ref{tab:table_2}.

Despite the reduced optimization headroom available in the block-wise setting, $\mH_\mZ$-PTQ consistently achieves the lowest perplexity and highest accuracy across all evaluated architectures and quantization formats. The improvement persists in both W4 and W4A4 settings, indicating that the proposed objective benefits not only transformation design, but also downstream range selection and rounding decisions.
Notably, the relative gains remain stable across model families, further supporting the robustness of the proposed framework.

\subsection{Ablation Studies}
\paragraph{Contribution of Individual Components} Table~\ref{tab:component_ablation} evaluates the contribution of each component within $\mH_\mZ$-PTQ in the W4A4KV4 setting. 
Removing any component results in a measurable degradation in performance, demonstrating that the final gains arise from the interaction of all optimization stages rather than from a single dominant factor. Among the examined components, the transformation stage provides the largest individual contribution, followed by activation clipping and Hessian modeling. 
Replacing the proposed output Kronecker Hessian approximation with a simpler gradient-free Hessian estimate also degrades performance, providing evidence that the improved results are not solely due to the optimization procedure but also depend on the quality of the underlying curvature model.
\begin{table}[!t]
    \centering
    \caption{Ablation for the components for $\mH_\mZ$-PTQ on W4A4KV4.  $-$Round refers to a version with round-to-nearest and a standard $L_p$ weight range-setting. $-$Trans. refers to the use of Hadamard transform instead of our gradient alignment transforms.  $-$Hess. indicates the use of a GPTQ-style hessian approximation instead of our output Kronecker approximation.}
    \label{tab:component_ablation}
    \footnotesize
    \setlength{\tabcolsep}{2.4pt}
    \begin{tabular}{l|ll|ll|ll|ll} \toprule & \multicolumn{2}{c|}{\textit{Llama v3.2 1B inst.}} & \multicolumn{2}{c|}{\textit{Llama v3 8B}} & \multicolumn{2}{c|}{\textit{Ministral 8B inst.}} & \multicolumn{2}{c}{\textit{Qwen v3 8B}} \\ Method & wiki ($\downarrow$) & csr ($\uparrow$) & wiki ($\downarrow$) & csr ($\uparrow$)& wiki ($\downarrow$) & csr ($\uparrow$)& wiki ($\downarrow$) & csr ($\uparrow$)\\ \midrule 
$\mH_\mZ$-PTQ & ${16.05}^{\pm 0.02}$ & ${54.11}^{\pm 0.19}$ & ${7.06}^{\pm 0.01}$ & ${67.25}^{\pm 0.23}$ & ${7.69}^{\pm 0.01}$ & ${69.30}^{\pm 0.26}$ & ${10.43}^{\pm 0.03}$ & ${65.56}^{\pm 0.46}$  \\\midrule
~~$-$Round & ${16.52}^{\pm 0.07}$ & ${53.13}^{\pm 0.57}$ & ${7.13}^{\pm 0.00}$ & ${66.54}^{\pm 0.04}$ & ${7.74}^{\pm 0.02}$ & ${69.10}^{\pm 0.15}$ & ${10.56}^{\pm 0.04}$ & ${65.14}^{\pm 0.42}$ \\
~~$-$A.clip & ${17.05}^{\pm 0.03}$ & ${52.35}^{\pm 0.36}$ & ${7.41}^{\pm 0.01}$ & ${65.72}^{\pm 0.14}$ & ${7.89}^{\pm 0.01}$ & ${68.35}^{\pm 0.62}$ & ${10.63}^{\pm 0.05}$ & ${64.08}^{\pm 0.05}$ \\
~~$-$Tran. & ${18.64}^{\pm 0.13}$ & ${51.68}^{\pm 0.86}$ & ${7.61}^{\pm 0.01}$ & ${66.06}^{\pm 0.39}$ & ${8.13}^{\pm 0.01}$ & ${67.98}^{\pm 0.30}$ & ${25.19}^{\pm 0.90}$ & ${55.80}^{\pm 0.48}$ \\
~~$-$Hess. & ${16.83}^{\pm 0.10}$ & ${52.80}^{\pm 0.64}$ & ${7.37}^{\pm 0.02}$ & ${65.75}^{\pm 0.11}$ & ${7.83}^{\pm 0.03}$ & ${68.90}^{\pm 0.34}$ & ${10.42}^{\pm 0.04}$ & ${65.77}^{\pm 0.03}$ \\
% $\mH_\mZ$-PTQ & ${54.11}^{\pm 0.19}$ & ${67.25}^{\pm 0.23}$ & ${69.30}^{\pm 0.26}$ & ${65.56}^{\pm 0.46}$ \\
% ~~$-$Round & ${53.13}^{\pm 0.57}$ & ${66.54}^{\pm 0.04}$ & ${69.10}^{\pm 0.15}$ & ${65.14}^{\pm 0.42}$ \\
% ~~$-$Act. clip & ${52.48}^{\pm 0.39}$ & ${65.65}^{\pm 0.10}$ & ${68.35}^{\pm 0.62}$ & 64.14 \\
% ~~$-$Transform & ${51.77}^{\pm 1.19}$ & ${66.06}^{\pm 0.39}$ & ${67.98}^{\pm 0.30}$ & NaN \\
% ~~$-$Hessian & ${52.43}^{\pm 0.16}$ & ${65.72}^{\pm 0.13}$ & ${68.84}^{\pm 0.45}$ & ${65.79}^{\pm 0.02}$ \\
\cline{1-9}
\bottomrule
\end{tabular}
    
\end{table}
\paragraph{Hessian-Guided Weight Range Selection}
\begin{figure}
    \centering
    \includegraphics[width=0.9\linewidth]{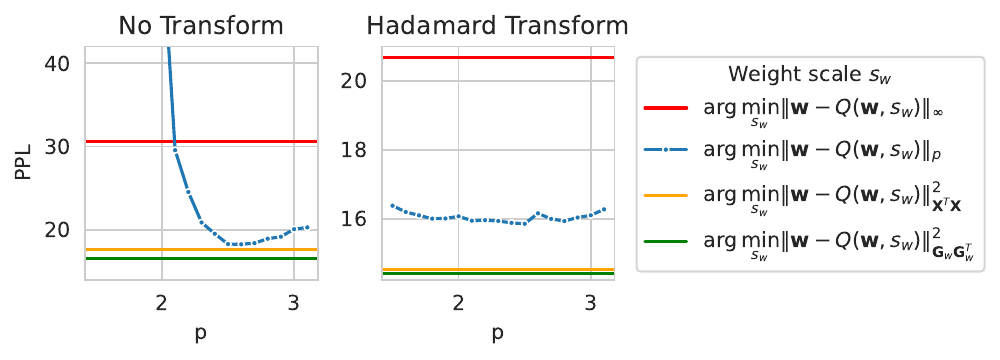}
    \vspace{-0.5cm}
    \caption{WikiText perplexity obtained by several weight scale-setting objectives on Llama 3.2 1B inst. under W4 per-channel quantization, with and without Hadamard transformations. 
    % Reconstruction-based methods require selecting an appropriate $L_p$ objective, and the optimal value of $p$ varies across settings. In contrast, 
    Weighting the quantization error using the proposed Fisher estimate consistently exceeds the best reconstruction objective without requiring hyperparameter tuning.}
    \label{fig:weight_ranges}
\end{figure}
Existing range-setting methods typically minimize an $L_p$ reconstruction objective, requiring the practitioner to select an appropriate value of $p$. Figure~\ref{fig:weight_ranges} evaluates the sensitivity of this choice. 
The optimal value of $p$ varies substantially across quantization settings and transformation choices, making fixed reconstruction objectives difficult to tune. In contrast, the proposed Hessian-weighted objective proposed in eq. (\ref{weight_range_eq}) directly incorporates downstream sensitivity information and consistently matches or outperforms the best manually selected $L_p$ objective without requiring any hyperparameter tuning.
These results suggest that incorporating curvature information into scale selection provides a more reliable objective than purely reconstruction-based metrics.

\paragraph{Hessian estimation and Rounding Strategies} 
\begin{table}[!t]
    \centering
    \caption{Comparison of rounding algorithms for different hessian estimates on the \textit{Llama v3.2 1B instruct} architecture in two quantization settings. The bottom rows report the effect of applying a Hadamard transform prior to quantization. ${\bf G}_W{\bf G}_W^T \otimes {\bf G}_Z^T{\bf G}_Z$ refers to our approximation. }
    \footnotesize
    \setlength{\tabcolsep}{3pt}
    \begin{tabular}{cc|ccc|ccc} \toprule & & \multicolumn{3}{c|}{W4 Per-channel} & \multicolumn{3}{c}{W4 Per-block (32)} \\ Tran. & Hessian Estimate & GPTQ & YAQA & \bf{ours} & GPTQ & YAQA & \bf{ours} \\ \midrule \multirow[m]{3}{*}{Id.} & ${\bf X}^T{\bf X} \otimes {\bf I}$ & \cellcolor{BlockColor}${30.96}^{\pm 4.58}$ & ${30.56}^{\pm 0.00}$ & ${17.65}^{\pm 0.09}$ & \cellcolor{BlockColor}${14.35}^{\pm 0.04}$ & ${69.59}^{\pm 10.14}$ & ${14.11}^{\pm 0.07}$ \\ & ${\bf G}_W{\bf G}_W^T \otimes {\bf G}_W^T{\bf G}_W$ & ${18.42}^{\pm 0.48}$ & \cellcolor{BlockColor}${18.81}^{\pm 0.10}$ & ${15.34}^{\pm 0.03}$ & ${14.28}^{\pm 0.11}$ & \cellcolor{BlockColor}${13.75}^{\pm 0.05}$ & ${\bf 13.64}^{\pm 0.05}$ \\ & ${\bf G}_W{\bf G}_W^T \otimes {\bf G}_Z^T{\bf G}_Z$ & ${19.66}^{\pm 3.89}$ & ${19.49}^{\pm 0.42}$ & \cellcolor{BlockColor}${\bf 15.23}^{\pm 0.06}$ & ${14.26}^{\pm 0.25}$ & ${13.92}^{\pm 0.08}$ & \cellcolor{BlockColor}${\bf 13.64}^{\pm 0.02}$ \\ \midrule \multirow[m]{3}{*}{Had.} & ${\bf X}^T{\bf X} \otimes {\bf I}$ & \cellcolor{BlockColor}${14.96}^{\pm 0.08}$ & ${20.68}^{\pm 0.00}$ & ${14.50}^{\pm 0.01}$ & \cellcolor{BlockColor}${13.68}^{\pm 0.03}$ & ${13.90}^{\pm 0.07}$ & ${13.69}^{\pm 0.05}$ \\ & ${\bf G}_W{\bf G}_W^T \otimes {\bf G}_W^T{\bf G}_W$ & ${14.50}^{\pm 0.06}$ & \cellcolor{BlockColor}${19.12}^{\pm 0.29}$ & ${14.26}^{\pm 0.05}$ & ${13.61}^{\pm 0.01}$ & \cellcolor{BlockColor}${13.70}^{\pm 0.05}$ & ${\bf 13.56}^{\pm 0.02}$ \\ & ${\bf G}_W{\bf G}_W^T \otimes {\bf G}_Z^T{\bf G}_Z$ & ${14.49}^{\pm 0.04}$ & ${19.29}^{\pm 0.24}$ & \cellcolor{BlockColor}${\bf 14.25}^{\pm 0.08}$ & ${13.60}^{\pm 0.02}$ & ${13.66}^{\pm 0.01}$ & \cellcolor{BlockColor}${\bf 13.56}^{\pm 0.01}$ \\ \cline{1-8} \bottomrule \end{tabular}
    \label{tab:rounding}
\end{table}
Table~\ref{tab:rounding} 
compares different rounding algorithms under several Hessian approximations. This experiment disentangles the effect of the curvature estimate from that of the rounding procedure itself.
First, for a fixed rounding algorithm, our proposed output Kronecker Hessian proxy performs on par with or outperforms the YAQA-style estimate, outperforming the gradient-free GPTQ-style hessian sketches.

Second, for a fixed Hessian estimate, incorporating our Hessian-aware rounding and range-setting procedure consistently improves performance over conventional GPTQ/LDLQ and YAQA optimization. This observation suggests that using hessian information to set the weight ranges while performing the rounding correction can better incorporate curvature information into the local decisions. This is especially relevant for coarse grained  quantization granularities.
Note that when using GPTQ rounding, the output Kronecker and YAQA-style hessian are equivalent since only the first component of the Kronecker product determines the weight corrections.

\section{Discussion and Conclusion}
We introduced a unified Hessian-guided framework for post-training quantization. Unlike existing approaches that optimize transformations, range selection, activation clipping, and rounding independently and using separate reconstruction objectives, our framework derives all quantization decisions from a common approximation of downstream prediction loss. Central to this framework is the proposed Output Kronecker approximation, a tractable output-space Fisher factorization that better correlates with downstream KL divergence than conventional local reconstruction metrics.
Across diverse architectures and quantization granularities, jointly optimizing these components consistently improved performance over existing state-of-the-art methods. These results suggest that much of the PTQ design space can be viewed as a single downstream optimization problem rather than a collection of independent heuristics.

The method, nevertheless, remains an approximation to the ideal downstream objective. Its derivation relies on a second-order expansion, a Kronecker-rank-one Fisher approximation, a decomposition of activation and weight errors, and rounding-noise and covariance assumptions. Moreover, estimating output gradients adds calibration time and memory relative to gradient-free reconstruction methods. Future work could investigate richer output-curvature models, adaptive approximations that trade accuracy for calibration cost, and extensions to non-uniform, mixed-precision, and hardware-aware quantization.

\bibliography{iclr2027_conference}
\bibliographystyle{iclr2027_conference}

\newpage
\appendix
\section{Quantization loss decomposition}
\label{app:proofs}

\subsection{Derivation of the transformation objective}

\label{loss_decomposition_derivations}
We derive Eq.~\ref{loss:approx_full} from the output KL divergence, first separating activation and weight errors to make the derivation of the optimal transform tractable.

\paragraph{From output KL to a quadratic proxy.}
Using the second-order KL approximation in Eq.~\ref{loss:step_1} and the Output Kronecker approximation in Eq.~\ref{eq:output_kronecker}, we obtain
\begin{align}
    \E{\KL{p_{\mZ}}{p_{\quant{\mZ}}}}
    &\approx \frac{1}{2}\E{\|\Delta\mZ\|_{\mF_{\otimes\mZ}}^2}.
    \label{app:expected_kl_proxy}
\end{align}
For a label $y\sim p_{\mZ}$, define $\mG_{\mZ}=\nabla_{\mZ}[-\log p_{\mZ}(y)]$.
Here and below, expectations average calibration inputs, sampled labels, and the rounding-noise model defined next.

For the trace calculations, write $\mA=\mG_{\mZ}\mG_{\mZ}^T$, $\mB=\mG_{\mZ}^T\mG_{\mZ}$, and $c=\|\mG_{\mZ}\|_F^2$.
For $c>0$ and any output perturbation $\mU$, row vectorization gives $\|\mU\|_{\mF_{\otimes\mZ}}^2=\trace{\mA\mU\mB\mU^T}/c$.

\paragraph{Rounding model and error decomposition.}
For this derivation, activation row $t$ uses grid step $d_{x,t}$ and weight column $j$ uses grid step $d_{w,j}$.
We model scalar rounding errors as conditionally independent and uniform on their quantization bins, with negligible clipping.
Conditioning on the full-precision activations, weights, sampled gradients, and grid steps, the errors have zero mean and variances $d_{x,t}^2/12$ and $d_{w,j}^2/12$.

The output perturbation satisfies $\Delta\mZ=\mX\Delta\mW+\Delta\mX\mW-\Delta\mX\Delta\mW$.
The product $\Delta\mX\Delta\mW$ contributes terms of third and fourth order in the quadratic proxy, so we omit it in the second-order approximation.
Conditional independence and zero means eliminate the mixed quadratic term between $\mX\Delta\mW$ and $\Delta\mX\mW$.
This gives the activation--weight decomposition in Eq.~\ref{loss:step_2}, following \citet{Federici2026}.

\paragraph{Activation contribution.}
The chain rule gives $\mG_{\mX}=\mG_{\mZ}\mW^T$.
Using the trace form of the proxy,
\begin{align}
    \|\Delta\mX\mW\|_{\mF_{\otimes\mZ}}^2
    &=\frac{\trace{\Delta\mX^T\mA\Delta\mX\mG_{\mX}^T\mG_{\mX}}}{c}.
\end{align}
Only matching scalar error indices survive the conditional expectation. Thus
\begin{align}
    \mathbb{E}[\Delta\mX^T\mA\Delta\mX]
    &=\frac{1}{12}\left(\sum_t A_{tt}d_{x,t}^2\right)\eye_i,\nonumber\\
    \mathbb{E}[\|\Delta\mX\mW\|_{\mF_{\otimes\mZ}}^2]
    &=\frac{1}{12}\left(\sum_t\alpha_t d_{x,t}^2\right)\|\mG_{\mX}\|_F^2,
    \label{app:activation_noise_average}
\end{align}
using $\alpha_t=A_{tt}/c$.

Write $\vx_t=\mX_{t,:}$ and $\range{\vx_t}=(2^{b_x}-1)d_{x,t}$.
For our asymmetric per-token activation grid, $\range{\vx_t}=\max_a X_{ta}-\min_a X_{ta}$.

\paragraph{Weight contribution.}
For the weights, $\mG_{\mW}=\mX^T\mG_{\mZ}$.
The same calculation contracts $\Delta\mW\mB\Delta\mW^T$, whose conditional expectation is $\frac{1}{12}(\sum_j B_{jj}d_{w,j}^2)\eye_i$. Thus
\begin{align}
    \mathbb{E}[\|\mX\Delta\mW\|_{\mF_{\otimes\mZ}}^2]
    &=\frac{1}{12}\left(\sum_j\beta_j d_{w,j}^2\right)\|\mG_{\mW}\|_F^2,
    \label{app:weight_noise_average}
\end{align}
using $\beta_j=B_{jj}/c$.
For column $\vw_j=\mW_{:,j}$, define $\range{\vw_j}=(2^{b_w}-1)d_{w,j}$ using the step of the configured symmetric weight grid.
This range is the grid span, with the signed endpoints defined in Section~\ref{sec:background}.

\paragraph{Separating ranges and gradient norms.}
Let $R_x=\sum_t\alpha_t\range{\vx_t}^2$ and $R_w=\sum_j\beta_j\range{\vw_j}^2$.
Averaging the conditional results gives $\E{R_x\|\mG_{\mX}\|_F^2}/[12(2^{b_x}-1)^2]$ and $\E{R_w\|\mG_{\mW}\|_F^2}/[12(2^{b_w}-1)^2]$ for the activation and weight contributions, respectively. To split the range and gradient terms we assume $\Cov(R_x,\|\mG_{\mX}\|_F^2)\approx0$ and $\Cov(R_w,\|\mG_{\mW}\|_F^2)\approx0$ over calibration inputs and sampled labels.
Applying $\E{UV}=\E{U}\E{V}+\Cov(U,V)$ to the two contributions yields
\begin{align}
    \E{\|\Delta\mX\mW\|_{\mF_{\otimes\mZ}}^2}
    &\approx\frac{\E{\sum_t\alpha_t\range{\vx_t}^2}}{12(2^{b_x}-1)^2}\E{\|\mG_{\mX}\|_F^2},
    \label{DX_to_GX}\\
    \E{\|\mX\Delta\mW\|_{\mF_{\otimes\mZ}}^2}
    &\approx\frac{\E{\sum_j\beta_j\range{\vw_j}^2}}{12(2^{b_w}-1)^2}\E{\|\mG_{\mW}\|_F^2}.
    \label{DW_to_GW}
\end{align}

Combining these with Eq. \ref{loss:step_2}  gives Eq. \ref{loss:approx_full}.

\subsection{Geometric-mean solution for gradient balancing}
\label{app:gradient_balance}

We derive the solution of the gradient-norm objective in Eq.~\ref{eq:gradient_balance}.
Define the gradient second-moment matrices
\begin{align}
    \mA &= \E{\mG_{\mW}\mG_{\mW}^T},
    &\mB &= \E{\mG_{\mX}^T\mG_{\mX}},
\end{align}
and assume $\mA,\mB\succ0$.
For any invertible $\mM$, set $\mP=\mM\mM^T\succ0$.
Using $\|\mU\|_F^2=\trace{\mU^T\mU}$ and cyclicity of the trace, the two terms become
\begin{align}
    \E{\|\mG_{\mX}\mM^{-T}\|_F^2}
    &= \trace{\mM^{-1}\mB\mM^{-T}}
     = \trace{\mB\mP^{-1}},\nonumber\\
    \E{\|\mM^T\mG_{\mW}\|_F^2}
    &= \trace{\mM^T\mA\mM}
     = \trace{\mA\mP}.
\end{align}
The objective therefore depends on $\mM$ only through $\mP$.
Every positive-definite $\mP$ has a unique symmetric positive-definite square root $\mM=\mP^{1/2}$, so Eq.~\ref{eq:gradient_balance} is equivalent to
\begin{align}
    \min_{\mP\succ0}\;\trace{\mA\mP}+\trace{\mB\mP^{-1}}.
    \label{app:gradient_trace_minimization}
\end{align}

This is exactly the trace minimization in \citet[Section~2.2, Eq.~(8)]{Zadeh2016}.
Their solution following Eq.~(12) gives the unique minimizer
\begin{align}
    \mP^*
    &= \mA^{-1}\#\mB
     = \mA^{-1/2}\left(\mA^{1/2}\mB\mA^{1/2}\right)^{1/2}\mA^{-1/2},
\end{align}
where $\#$ is the Pusz--Woronowicz geometric mean \citep{pusz1975}.
Taking the symmetric positive-definite square root yields
\begin{align}
    \mM^*=(\mA^{-1}\#\mB)^{1/2},
\end{align}
which is the solution stated in Eq.~\ref{eq:gradient_balance}.
Over all invertible matrices, the minimizers are $\mM=(\mP^*)^{1/2}\mR$ with $\mR\mR^T=\eye_i$.
Thus the gradient-norm objective fixes the positive-definite component while leaving the rotation free for the subsequent range-reduction step.

\subsection{Range clipping for weights derivation}
Our algorithm applies the selected transformation before setting quantization ranges and rounding the weights. These steps require gradients and Fisher estimates in the transformed coordinates. The chain rule lets us compute these quantities from the original gradient estimates, avoiding an additional backward pass through the model.
For the transformed coordinates defined in Section~\ref{sec:background}, the chain rule gives
\begin{align}
    \mG_{\mX_{\mT}}&=\mG_{\mZ}\mW_{\mT}^T=\mG_{\mX}\mT^{-T},
    &\mG_{\mW_{\mT}}&=\mX_{\mT}^T\mG_{\mZ}=\mT^T\mG_{\mW}.
    \label{app:transformed_gradients}
\end{align}
This shows that, instead of estimating the Fisher approximation directly from $\mG_{\mW_{\mT}}$, we can transform the existing approximation computed from the untransformed gradient $\mG_{\mW}$, using the identity $\mG_{\mW_{\mT}}=\mT^T\mG_{\mW}$. This relation is used in Eq.~\ref{weight_range_eq} and in the weight contribution to Eq.~\ref{loss:approx_full}.
%Apply the same noise and covariance assumptions in these coordinates, with grid ranges computed from the transformed rows and columns.
%Substituting these identities into Eq. \ref{DW_to_GW} and Eq. \ref{DX_to_GX} gives Eq.~\ref{loss:approx_full}.

For fixed $\mT$ and $\Delta\mW_{\mT}$, write
$\mP=\mG_{\mW_{\mT}}\mG_{\mW_{\mT}}^T$ and
$\mQ=\mG_{\mZ}^T\mG_{\mZ}/\|\mG_{\mZ}\|_F^2$.
We assume that $\|\mG_{\mZ}\|_F^2$ is constant, and that the correlations between the input- and output-side curvature factors contribute negligibly to the expected weight-error proxy:
\begin{align}
    \E{\trace{
        (\mP-\E{\mP})\Delta\mW_{\mT}
        (\mQ-\E{\mQ})\Delta\mW_{\mT}^T
    }} &\approx 0.
    \label{app:weight_range_covariance}
\end{align}
Using $\mG_{\mW_{\mT}}=\mX_{\mT}^T\mG_{\mZ}$ in the
trace form of the proxy gives
\begin{align}
    \E{\|\mX_{\mT}\Delta\mW_{\mT}\|_{\mF_{\otimes\mZ}}^2}
    &= \E{\trace{
        \Delta\mW_{\mT}\mQ\Delta\mW_{\mT}^T\mP
    }} \nonumber\\
    &\approx \trace{
        \Delta\mW_{\mT}\E{\mQ}\Delta\mW_{\mT}^T\E{\mP}
    }.
    \label{app:weight_range_factorization}
\end{align}
Under the constant-norm assumption,
$\E{\mQ}=\E{\mG_{\mZ}^T\mG_{\mZ}}/\E{\|\mG_{\mZ}\|_F^2}$.
Substituting the definitions of $\mP$ and $\mQ$ and applying
row vectorization therefore yields
\begin{align}
    \E{\|\mX_{\mT}\Delta\mW_{\mT}\|_{\mF_{\otimes\mZ}}^2}
    &\approx
    \|\Delta\mW_{\mT}\|_{
        \frac{
            \E{\mG_{\mW_{\mT}}\mG_{\mW_{\mT}}^T}
            \otimes
            \E{\mG_{\mZ}^T\mG_{\mZ}}
        }{\E{\|\mG_{\mZ}\|_F^2}}
    }^2.
\end{align}
\section{Experimental details}
\label{app:experimental_details}

\subsection{Models and Baselines} We evaluate across multiple model families and scales, including \textit{Llama v3 8B} and \textit{Llama v3.2 1B instruct} \citep{grattafiori_llama_2024}, \textit{Ministral 8B instruct} \citep{ministral}, and \textit{Qwen v3 8B} \citep{qwen3}. We compare against representative state-of-the-art transformation methods, including Quarot \citep{ashkboos_quarot_2024}, FlatQuant \citep{sun_flatquant_2024}, CAT \citep{Federici2026} and WUSH \citep{chen2026wush}. For rounding algorithms, we consider GPTQ/LDLQ  \citep{frantar_gptq_2022, chee_quip_2023}, and YAQA Sketch B \citep{Tseng25}.

\subsection{Detailed Setup}
The experimental configuration for the results reported in Table~\ref{tab:table_1} and Table~\ref{tab:table_2} is reported in Table~\ref{tab:exp_details_1} and Table~\ref{tab:exp_details_2}, respectively. In the tables $\boldsymbol{\Sigma}_\mX$ refers to $\E{\mX^T\mX}$, $\boldsymbol{\Sigma}_\mW$ to $\mW\mW^T$, and $\boldsymbol{\Sigma}_{\mG_\mX}=\E{\mG_\mX^T\mG_\mX}$, $\boldsymbol{\Sigma}_{\mG_\mW}=\E{\mG_\mW\mG_\mW^T}$.
The matrix $\mU$ refers to the eigen-decomposition of $\boldsymbol{\Sigma}_\mX^{-1}\#\boldsymbol{\Sigma}_\mW=\mU\Lambda\mU^T$. $B(\cdot)$ refers to the extraction of a block-diagonal component. Note that for block-wise quantization experiments, we
omit the activation clipping strategy because of the limited impact on the small block size.

For all the experiments, we use 500 candidates sampled uniformly in $[\max_i|\vw_i|, 2\max_i|\vw_i|]$ to determine the minimizer for Hessian-weighted and $L_p$ norms. For the minimization of the clipping parameter, we use only 100 candidates uniformly sampled in $[0.5,1]$, determining the best by summing the error over 32 different sequences.
In block-wise quantization settings, both YAQA and our algorithm process 32 output channels at a time during rounding. We select the quantization scale independently for each block, but batch the scale search across the 32 blocks. This reduces runtime without significantly affecting quantization performance.

\begin{table}[!t]
    \centering
    \caption{Details on the setting used to produce Table~\ref{tab:table_1}. The block size for the transforms used for CAT and our method is 128.}
    \label{tab:exp_details_1}
    \footnotesize
    \begin{tabular}{l|c|c|c|c}
        \toprule
         Method & W. Scale & W. Round & Act. Clip & Transforms\\
         \midrule
         RTN & $2\max_i|\vw_i|$ & None & $\gamma=1$  & None\\
         QuaRot$^{+GPTQ}$ & $\min_{s_w}\|\vw-Q(\vw;s_w)\|^{2.4}$ & GPTQ & $\gamma=1$  & (Fused) $\mH$\\
         CAT$^{+GPTQ}$ & $\min_{s_w}\|\vw-Q(\vw;s_w)\|^{2.4}$ & GPTQ & $\gamma=1$  & $B(\boldsymbol{\Sigma}_\mX^{-1}\#\boldsymbol{\Sigma}_\mW)^{1/2}\mH$\\
         FlatQuant$^{+GPTQ}$ & Learned $s_w$ & GPTQ & Learned $\gamma$ & Learned $\mT_i \otimes \mT_o$\\
         $\mH_\mZ$-PTQ & \multicolumn{2}{c|}{Algorithm in Sec.~\ref{sec:range_and_round}} &  Sec.~\ref{sec:clip} & $B(\boldsymbol{\Sigma}_{\mG_\mW}^{-1}\#\boldsymbol{\Sigma}_{\mG_\mX})^{1/2}\mH$\\
         \cline{1-5}
         \bottomrule
         
    \end{tabular}
    
\end{table}

\begin{table}[!t]
    \centering
    \caption{Details on the setting used to produce Table~\ref{tab:table_2}. The transformation in this table already refer to a single block of size 32.}
    \label{tab:exp_details_2}
    \footnotesize
    \begin{tabular}{l|c|c|c|c}
        \toprule
         Method & Weight Scale & Weight Round & Act. Clip & Block Transforms\\
         \midrule
         RTN & $2\max_i|\vw_i|$ & None & $\gamma=1$  & None\\
         Had.$^{+GPTQ}$ & $2\max_i|\vw_i|$ & GPTQ & $\gamma=1$  & Hadamard\\
         Had.$^{+YAQA}$ & $2\max_i|\vw_i|$ & YAQA & $\gamma=1$  & Hadamard\\
         CAT$^{+GPTQ}$ & $2\max_i|\vw_i|$ & GPTQ & $\gamma=1$  & $(\boldsymbol{\Sigma}_\mX^{-1}\#\boldsymbol{\Sigma}_\mW)^{1/2}\mH$ \\
         WUSH$^{+GPTQ}$ & $2\max_i|\vw_i|$ & GPTQ & $\gamma=1$ & $(\boldsymbol{\Sigma}_\mX^{-1}\#\boldsymbol{\Sigma}_\mW)^{1/2}\mU\mH$\\
         $\mH_\mZ$-PTQ & \multicolumn{2}{c|}{Algorithm in Sec.~\ref{sec:range_and_round}} &  $\gamma=1$  & $(\boldsymbol{\Sigma}_{\mG_\mW}^{-1}\#\boldsymbol{\Sigma}_{\mG_\mX})^{1/2}\mH$\\
         \cline{1-5}
         \bottomrule
         
    \end{tabular}
    
\end{table}
\section{Additional Results}
\label{app:additional_results}

\begin{figure}[!t]
    \centering
    \includegraphics[width=\linewidth]{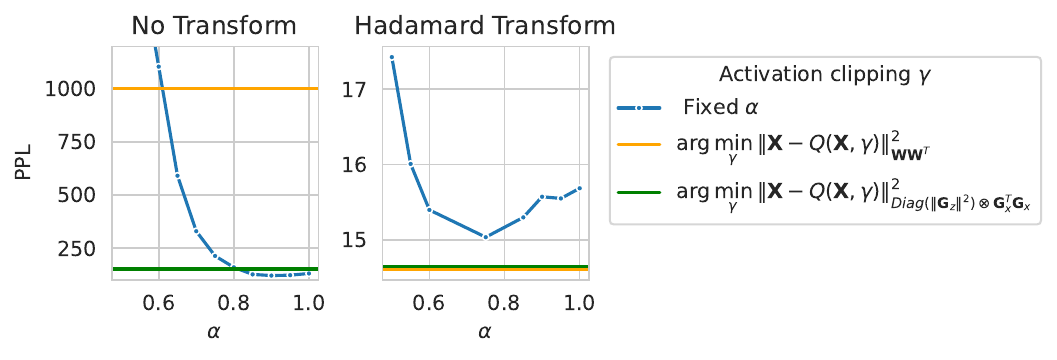}
    \caption{Wikitext perplexity obtained by different activation clipping objectives on Llama 3.2 1B Instruct under 4-bit per token quantization, with and without Hadamard transformations. }
    \label{fig:act_clipping}
\end{figure}

\paragraph{Activation Clipping}
Figure~\ref{fig:act_clipping} shows the effect of different activation clipping strategies. Overall, using the proposed Hessian proxy $\E{\diag{[\|\vg_{\mZ,1}\|^2_2,\hdots,\|\vg_{\mZ,s}\|^2_2]}}\otimes\E{\mG_\mX^T\mG_\mX}$, performs strongly against manual selection of the clipping percentage or clipping using simpler hessian proxies. However, the sub-optimality is likely due to the fact that the same "average" offline estimated curvature is used for all tokens instead of using a curvature estimation for each token independently. For $\mH_{\mZ}$-PTQ we assume the covariances have been transformed by T according to Eq. \ref{app:transformed_gradients}.

\paragraph{Results with simple Round to Nearest}
Table~\ref{tab:table_1_no_round} reports the results for the methods and quantization setting in Table~\ref{tab:table_1} without any weight rounding adjustment. Our $\mH_\mZ$-PTQ still outperforms the baselines even without the improved weight-rounding and scale setting algorithm, demonstrating the effectiveness of our transformations and activation clipping method.
\begin{table}[!th]
    \centering
    \caption{A4W4KV4 Quantization results without any weight rounding optimization and standard $L_p$-based ranges.}
    \label{tab:table_1_no_round}
    \setlength{\tabcolsep}{1.5pt}
    \scriptsize
    \begin{tabular}{l|ll|ll|ll|ll} \toprule & \multicolumn{2}{c|}{\textit{Llama v3.2 1B inst.}} & \multicolumn{2}{c|}{\textit{Llama v3 8B}} & \multicolumn{2}{c|}{\textit{Ministral 8B inst.}} & \multicolumn{2}{c}{\textit{Qwen v3 8B}} \\ Method & wiki ($\downarrow$) & csr ($\uparrow$) & wiki ($\downarrow$) & csr ($\uparrow$)& wiki ($\downarrow$) & csr ($\uparrow$)& wiki ($\downarrow$) & csr ($\uparrow$)\\ \midrule 
FP & 13.16 & 57.61 & 6.14 & 70.12 & 6.97 & 71.29 & 9.73 & 67.46 \\\midrule
QuaRot & 29.77 & 46.23 & 10.99 & 57.07 & 9.14 & 64.55 & 13.68 & 57.33 \\
CAT & ${18.20}^{\pm 0.10}$ & ${51.41}^{\pm 0.28}$ & ${7.67}^{\pm 0.01}$ & ${63.90}^{\pm 0.24}$ & ${7.99}^{\pm 0.01}$ & ${67.72}^{\pm 0.20}$ & ${11.01}^{\pm 0.04}$ & ${63.66}^{\pm 0.05}$ \\
FlatQuant & ${17.09}^{\pm 0.04}$ & ${52.60}^{\pm 0.42}$ & ${7.50}^{\pm 0.52}$ & ${66.28}^{\pm 2.04}$ & ${7.84}^{\pm 0.05}$ & ${68.71}^{\pm 0.14}$ & ${10.55}^{\pm 0.13}$ & ${65.34}^{\pm 0.32}$ \\
$\mH_\mZ$-PTQ$^{-a.clip}$ & ${17.78}^{\pm 0.12}$ & ${51.42}^{\pm 0.16}$ & ${7.51}^{\pm 0.01}$ & ${64.55}^{\pm 0.56}$ & ${7.95}^{\pm 0.02}$ & ${68.23}^{\pm 0.25}$ & ${10.78}^{\pm 0.04}$ & ${63.21}^{\pm 0.00}$ \\
$\mH_\mZ$-PTQ & ${16.52}^{\pm 0.07}$ & ${53.13}^{\pm 0.57}$ & ${7.13}^{\pm 0.00}$ & ${66.54}^{\pm 0.04}$ & ${7.74}^{\pm 0.02}$ & ${69.10}^{\pm 0.15}$ & ${10.56}^{\pm 0.04}$ & ${65.14}^{\pm 0.42}$ \\
\bottomrule
\end{tabular}
    
\end{table}

\end{document}